\documentclass[letterpaper, 10 pt, conference]{ieeeconf}  % Comment this line out if you need a4paper

\usepackage[pdftex,dvipsnames]{xcolor}  % Coloured text etc.
\usepackage[colorinlistoftodos,prependcaption,textsize=tiny]{todonotes}

\IEEEoverridecommandlockouts                              % This command is only needed if you want to use the \thanks command
\usepackage{amsthm}
\usepackage{graphics} % for pdf, bitmapped graphics files
\usepackage{epsfig} % for postscript graphics files
\usepackage{times} % assumes new font selection scheme installed
\usepackage{amsmath} % assumes amsmath package installed
\usepackage{amssymb}  % assumes amsmath package installed
\usepackage{textcomp}

\usepackage{bm}
\usepackage{xcolor}
\usepackage{graphicx}
\usepackage{booktabs}
\usepackage{siunitx}% An awesome package for typesetting and manipulation numbers and units.
\usepackage{hyperref}% Used for insert the link
\usepackage{courier}% courier font
\usepackage{caption}% Better control over caption
\usepackage{lipsum}% Example text
\usepackage{multirow}
\usepackage{diagbox} 
\usepackage{subcaption}
\usepackage{amsfonts}
\usepackage[ruled,linesnumbered]{algorithm2e}
\usepackage[noend]{algpseudocode}
\usepackage{textcomp}
\usepackage{stfloats}
\usepackage{url}
\usepackage{verbatim}
\usepackage{graphicx}
\usepackage{times} % assumes new font selection scheme installed
\usepackage[maxbibnames=99,giveninits=true]{biblatex}
\usepackage{soul}

\usepackage{array}
\usepackage{arydshln}
\newcommand{\R}{\mathbb{R}}%commands for easy math notations

\newcommand{\bp}{{\bm p}_t}
\newcommand{\by}{{\bm y}_t}
\newcommand{\bz}{{\bm z}_t}

\newcommand{\ba}{\bm{\alpha}_t}
\newcommand{\Z}{\mathbb{Z}}

\newtheorem{definition}{\bf Definition}

\newtheorem{theorem}{\bf Theorem}
\newtheorem{lemma}{\bf Lemma}

\usepackage[normalem]{ulem}

\begin{document}
\title{\LARGE \bf Symbolic Perception Risk in Autonomous Driving  
}

\author{Guangyi Liu, Disha Kamale, Cristian-Ioan Vasile, and Nader Motee 
\thanks{
This work was supported by the ONR N00014-19-1-2478. \endgraf
G. Liu, D. Kamale, C.I. Vasile and N. Motee are with the Department of Mechanical Engineering and Mechanics, Lehigh University, Bethlehem, PA, 18015, USA. {\tt\small \{gliu,ddk320,cvasile,motee\}@lehigh.edu}.
}
}

\maketitle

% Replace this two for submission
\thispagestyle{plain}
\pagestyle{plain}
\newtheorem{problem}{Problem}[section]
\newcommand{\AP}{{AP}}
\newcommand{\TS}{\mathcal{T}}
\newcommand{\BB}[1]{\mathbb{#1}}
\newcommand{\BF}[1]{\mathbf{#1}}
\newcommand{\CA}[1]{\mathcal{#1}}
\newcommand{\card}[1]{\left| {#1} \right|}
\newcommand{\spow}[1]{2^{#1}}
% \thispagestyle{empty}
% \pagestyle{empty}

%%%%%%%%%%%%%%%%%%%%%%%%%%%%%%%%%%%%%%%%%%%%%%%%%%%%%%%%%%%%%%%%%%%%%%%%%%%%%%%%%%%%%
\begin{abstract}
We develop a novel framework to assess the risk of misperception in a traffic sign classification task in the presence of exogenous noise. We consider the problem in an autonomous driving setting, where detection accuracy gradually improves as the distance to traffic signs decreases due to improved resolution and smaller impact from noise.
The common accuracy measures for classification often do not reveal the severity of the potential cost from the misperception. Thus, for the estimated perception statistics obtained using the standard classification algorithms, we aim to quantify the risk of misperception to mitigate the effects of inaccurate detection.
By exploring perception outputs, their expected high-level actions, and potential costs, we show the closed-form representation of the conditional value-at-risk (CVaR) of misperception.
Moreover, we propose a discounted accumulated CVaR-based risk that leverages the increasing detection quality.
Several case studies support the effectiveness of our proposed methodology.  

% The Voronoi partitions of the belief space explicitly define a family of misperception sets, and the probability of misperception is computed using the exceedance probability of the estimated Dirichlet distribution of belief outputs. 

% \cristi{What about the accumulated version of CVaR? It connects to the incremental setting. Please check the above sentences that I added.}

% \cristi{The abstract does not describe the features of the detection modality considered which is at the heart of the work. Describe its incremental nature, and impact it can have.}

\end{abstract}

%%%%%%%%%%%%%%%%%%%%%%%%%%%%%%%%%%%%%%%%%%%%%%%%%%%%%%%%%%%%%%%%%%%%%%%%%%%%%%%%%%%%%
\section{Introduction}
``All humans are prone to make mistakes," 
which are especially crucial while performing safety-critical tasks such as driving. Given the fact that nearly $94\%$ of the accidents are caused by human error \cite{singh2015critical}, and more than $74\%$ among them are related to poor recognition and decision, the development of autonomous driving technologies has gained significant research attention in anticipation of improved human safety \cite{yurtsever2020survey}. However, due to the inevitable hardware limitations, algorithmic errors as well as external factors such as weather and illumination conditions, it is not rare to see autonomous vehicles performing imperfect recognition of the environment, surrounding vehicles, and making poor decisions that lead to undesirable consequences \cite{dixit2016autonomous,barabas2017current,endsley2017autonomous}. For instance, as presented in \cite{favaro2017examining}, most autonomous vehicle-related accidents are caused by poor recognition of the environment or surrounding vehicles. Therefore, to maintain the autonomous driving vehicle in a safe operating state in a noisy environment, one must assess the reliability of the noisy belief output.

\begin{figure}[t]
    \begin{subfigure}[t]{.49\linewidth}
        \centering
    	\includegraphics[width=\linewidth]{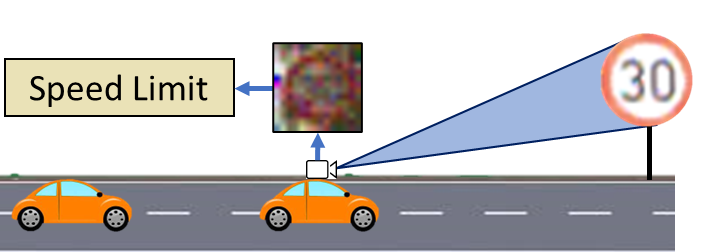}
    	\caption{The case when the belief output is {\it correct} and there is no accident.}
    \end{subfigure}
    \medskip
    \begin{subfigure}[t]{.49\linewidth}
        \centering
    	\includegraphics[width=\linewidth]{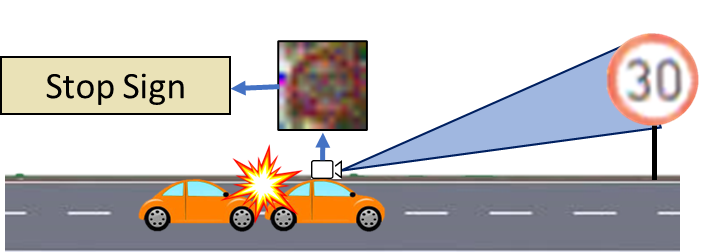}
    	\caption{The case when {\it misperception} occurs and its corresponding accident.}
    \end{subfigure}
	\caption{The above diagram depicts the possible outcomes of the misperception.}
 \vspace{-5pt}
% 	the scenario when autonomous vehicles driving towards the traffic sign and
    \label{fig:diag}
\end{figure}

We consider the motivational scenario when an autonomous vehicle is driving towards a traffic sign, as depicted in Fig. \ref{fig:diag}. The vehicle is equipped with an onboard camera, which detects, and aims to classify the traffic sign with one of the predefined labels. Due to the location and the unpredictable environment change, the detected image may suffer from various perturbations such as low resolution and pixel-wise noise, which reshape the belief output into a random variable. Our objective is to construct the notion of ``risk'' of misperception for a given perception model and its visual input, which can be merged into a planning \cite{kamale2022cautious} and decision-making module for minimizing the chance of systemic failures \cite{li2022learning, barbosa2021risk, hakobyan2019risk}.

In the past decades, some well-known risk measures, e.g., Value-at-Risk (VaR) \cite{rockafellar2000optimization} and Conditional Value-at-Risk (CVaR) \cite{rockafellar2002conditional}, have shown their significant advances in revealing the uncertainty and reliability of random variables given some harmful tail events. By treating the belief output as a random variable, we use the CVaR measure for the risk quantification, which evaluates the expected outcome when the system has entered the undesired state of operations, e.g., inter-vehicle accidents. The CVaR measure also reveals the severity \cite{wei2022cvar} of the failure when the undesired state is reached, which shows significant importance in our motivational scenario\footnote{Mis-perceiving the \textit{Speed-limit of 30 MPH} sign as a \textit{Stop} sign is more dangerous than as a \textit{Speed-limit of 15 MPH} sign.}. Consequently, it becomes essential to consider not only the chances of the potential systemic failures but also their magnitudes in terms of costs \cite{hruschka2019risk,majumdar2020should}.  

The risk quantification process begins with estimating statistics of the belief outputs with a Dirichlet distribution and using the concept of the Voronoi partitioning of the belief space \cite{liu2022robustness} to obtain the discrete distribution of the noisy belief output into class labels. Then, based on the user-defined cost metric for misperceiving each traffic sign, our main result evaluates the risk of traffic sign misperception regarding the severity of the potential accidents. 

Our distinct contributions with respect to the existing literature are multi-fold. First, owing to the coherence property, we adopt the Conditional Value-at-Risk (CVaR) measure and propose a risk quantification framework that evaluates the chances of misperception for a given noisy belief output. Secondly, the proposed framework can be customized with a user-defined cost metric and applied to most classification problems when the belief output statistics are available. Thirdly, the proposed approach is control agnostic; given the traffic sign label corresponding to the minimum risk value, our risk-quantification framework is amenable to any control approach. Finally, the case studies demonstrate a significantly improved performance in terms of safety when using the misperception risk to make decisions under varying levels of noise and resolution.

\section{Mathematical Notations}     

The $n$-dimensional Euclidean space with elements $\bm{z} = [z_1, \dots, z_n]^T$ is denoted by $\R^n$, and $\mathbb{R}_{+}$ will denote the positive orthant of $\R$. The collection of all integers is denoted by $\Z$, where $\Z_+$ will denote the positive orthant of $\Z$.
We represent the $n \times n$ identity matrix as $\bm{I}_{n}$ and the vector of all ones as $\bm{1}_n$, respectively. The dimension of a vector $\bm{z} \in \R^n$ is shown by $\dim(\bm{z}) = n$.
The $i$'th element of a vector $\bm{x}$ is shown by $x_{i}$, and when the vector is time indexed by $t$, the notation is adapted to $x_{t,i}$. The $(i,j)$'th entry of matrix $\bm{A}$ is represented by $A_{ij}$. 
Let us define the collection of all feasible belief vectors $\pi$, i.e., the belief space, as {$\mathcal{P}_n = \{\pi \in \R_{+}^{n} ~|~ \pi^T \bm{1}_n = 1 \}$}.
We also define the Gamma function as $\Gamma(z) = \int_0^{\infty} x^{z-1}e^{-x} dx$ for $\mathbf{Re}(z) > 0$, and the corresponding Digamma function as $\Psi(x) = \frac{\Gamma'(x)}{\Gamma(x)}$.

%%%%%%%%%%%%%%%%%%%%%%%%%%%%%%%%%%%%%%%%%%%%%%%%%%%%%%%%%%%%%%%%%%%%%%%%%%%%%%%%%%%%%
\section{Problem Statement} \label{sec:problem-statement}

In this paper, we tackle risk evaluation associated with (mis)perception of objects while an autonomous agent is in motion to perform its mission. Objects are semantically linked, and the agent's detection performance depends on its proximity to them.

For the remainder of the paper, let us consider the case of an autonomous vehicle equipped with an onboard camera traveling along a road with traffic signs with labels from $\mathcal{M} = \{1, \ldots, m\}$.
The car travels at a constant velocity towards a traffic sign that it will reach in $T \in \R_+$ time.
Within the time interval $[0, T]$, the car must take a high-level action from a given set $\Lambda$ to ensure conformance to traffic rules~\cite{vienna_conv}
based on a finite set of observations $\by$, $t \in [0, T]$,
that, ideally, matches the ground truth sign label $\ell \in \mathcal{M}$.
The observations $\by \in \R^{i_d}$ represent images $i_d = q_1 \times q_2$ from the onboard camera in vector form.

We model the detection process as a belief-valued function $h(\cdot)$ over the traffic labels $\mathcal{M}$. Formally, we have
$
    \bp = h(\by),
$
where the {\it belief output} $\bp \in \mathcal{P}_m$ is the output of the detection algorithm (see~\S\ref{sec:perception-model}).
The \emph{perception output} is the value of function $\arg\max\{ \bp \}\in \mathcal{M}$ for a given belief output $\bp \in \mathcal{P}_m$. The functionality of $\arg\max$ can be interpreted as one of the simplest forms of inference in the context of this work.  

In most real-world scenarios, the onboard cameras suffer from limited sensing range, resolutions, and noisy visual input.
As a result, the generated belief output can be potentially inaccurate and pertaining to noise.
The observation model of the image $\by \in \R^{i_d}$ is given by\footnote{The observation can also be taken from other closed-loop system dynamics, e.g., \cite{dean2020robust}.}
\begin{equation} \label{eq:modified_observation}
    \by = g(t, \bm{y}_0) + b_t \, \bm{\xi}_t,
\end{equation}
where $\bm{y}_0 \in \R^{i_d}$ denotes the high resolution and noise-free image of the traffic sign. These types of perturbation are commonly considered in the research of adversarial attacks of the image classification process using neural network models, see \cite{carrara2018detecting,dong2020benchmarking,subramanya2019fooling}.

The nonlinear function $g: [0,T] \times \R^{i_d} \rightarrow \R^{i_d}$ modifies the image $\bm{y}_0$ with various resolutions together with the exogenous disturbances $\bm{\xi}_t$, which denotes the vector of pixel-wise independent Brownian motions\footnote{One may use other types of noise to model effects of uncertainties.} with a time-varying diffusion coefficient $b_t$. The detail of this modification is further illustrated in \S \ref{sec:modified_gtsrb}. 

\begin{definition}
    For a given belief output $\bp \in \mathcal{P}_{m}$, a misperception occurs if 
    \begin{equation}    \label{eq:misperception}
        \arg\max \{\bp\} \neq \ell,
    \end{equation}
    where $\ell \in \mathcal{M}$ is the ground truth label of the sign.
\end{definition}

The {\it problem} is to quantify the risk of misperception as a function of the statistics of the noisy belief output $\bp$ and the given confidence level. To reveal the severity of the misperception, we incorporate the user-defined cost metric with the CVaR measure, which connects the event of misperception with the potential loss due to accidents. Once the risk is assessed, the perceived outcome, which corresponds to a traffic sign with the minimal risk level, can be used with any controller of choice, e.g., low-level controller, symbolic controller, or both~\cite{liu2017path,kamale2022cautious, schwarting2018planning}.

% The rest of the paper is organized as follows. First, in \S \ref{sec:preliminaries}, we present a few key preliminary results on the Voronoi partition of the belief space and the CVaR measure. Then, in \S \ref{sec:perception-model}, we show how the belief outputs can be estimated by a random variable that follows the Dirichlet distribution. These results help us shape the closed-form risk formula of misperception in \S \ref{sec:risk}, which constitutes the main contribution of this work. Finally, in \S \ref{sec:case}, by defining the available high-level actions, we validate the effectiveness of our proposed framework with extensive simulations.

%%%%%%%%%%%%%%%%%%%%%%%%%%%%%%%%%%%%%%%%%%%%%%%%%%%%%%%%%%%%%%%%%%%%%%%%%%%%%%%%%%%%%
\section{Preliminaries}     \label{sec:preliminaries}
For the exposition of our main result, let us first introduce some necessary results and definitions. 
\subsection{The Voronoi Partition of the Belief Space}
All possible belief outputs $\bp$ belong to the belief space $\mathcal{P}_m$, which is a $(m-1)-$simplex in $\R^m$. The perception model decides the label of the class is $i \in \mathcal{M}$ if and only if $p_{t,i} > p_{t,j}$ for all $j \in \mathcal{M}$ and $j \neq i$. The $\arg\max$ classification criterion can be explicitly represented via the Voronoi partitioning \cite{klein1988abstract}. The Voronoi partitions of the belief space $\mathcal{P}_m$ are given by $V_1, V_2, \dots, V_m$, where
\begin{equation}    \label{eq:v_k}
    V_i = \left\{\bp \in \mathcal{P}_m ~|~ p_{t,i} > p_{t,j}, ~  \forall j \neq i \right\}. 
\end{equation}
The above partitioning is equivalent to the $\arg\max$ criterion since $\arg \max \{ \bp \} = i$ if and only if $\bp \in V_i$, see \cite{liu2022robustness}.

\subsection{Conditional Value-at-Risk Measure}
To quantify the uncertainty level and the expected outcome encapsulated in belief outputs, we employ the notion of Conditional Value-at-Risk (CVaR) measure \cite{rockafellar2002conditional}. The CVaR indicates the severity of a random variable landing inside an undesirable set of values that characterizes the dangerous state of the system operation with specific confidence level, i.e., Value-at-Risk (VaR). In probability space $(\Omega, \mathcal{F}, \mathbb{P})$, the VaR of the random variable $Y: \Omega \rightarrow \R$ is defined as
\begin{equation}
    \mathcal{R}^{\varepsilon}_{VaR}(Y) = \min\{z \mid F_{Y} (z) \geq 1 - \varepsilon\},
\end{equation}
where the cumulative distribution function $F_Y(z) = \mathbb{P}\{Y \leq z\}$. Then, the CVaR is defined as follows.
\begin{definition}
    The Conditional-Value-at-Risk with the confidence level $(1-\varepsilon) \in [0,1]$ is the mean of the generalized $\varepsilon-$tail distribution:
\begin{equation} \label{eq:cvar}
    \mathcal{R}^{\varepsilon}_{CVaR} (Y) = \int_{-\infty}^{\infty} z ~ d F_{Y}^{\varepsilon}(z),
\end{equation}
where
\begin{equation}
    F_{Y}^{\varepsilon}(z) = 
    \begin{cases}
        0, \hspace{2.35cm} \text{if } z < \mathcal{R}^{\varepsilon}_{VaR}(Y)\\
        \frac{F_{Y}(z) + \varepsilon - 1}{\varepsilon}, \hspace{1cm} \text{if } z \geq \mathcal{R}^{\varepsilon}_{VaR}(Y)
    \end{cases}.
\end{equation}
\end{definition}
A smaller value of $\varepsilon$ indicates a higher level of confidence on random variable $Y$ to stay below  $\mathcal{R}^{\varepsilon}_{VaR}(Y)$. If $Y$ has a continuous distribution function, CVaR can be obtained as the conditional expectation of $Y$ subject to $Y \geq \mathcal{R}^{\varepsilon}_{VaR}(Y)$.
In the case of discrete distributions, one may need to split a probability atom, and CVaR may be obtained by averaging a fractional number of scenarios, see \cite{sarykalin2008value}.
% \guangyi{double check the cvar definition}

%%%%%%%%%%%%%%%%%%%%%%%%%%%%%%%%%%%%%%%%%%%%%%%%%%%%%%%%%%%%%%%%%%%%%%%%%%%%%%%%%%%%%
\section{Traffic Sign Perception Model}   \label{sec:model}
\begin{figure*}[t]
    \centering
	\includegraphics[width=\linewidth]{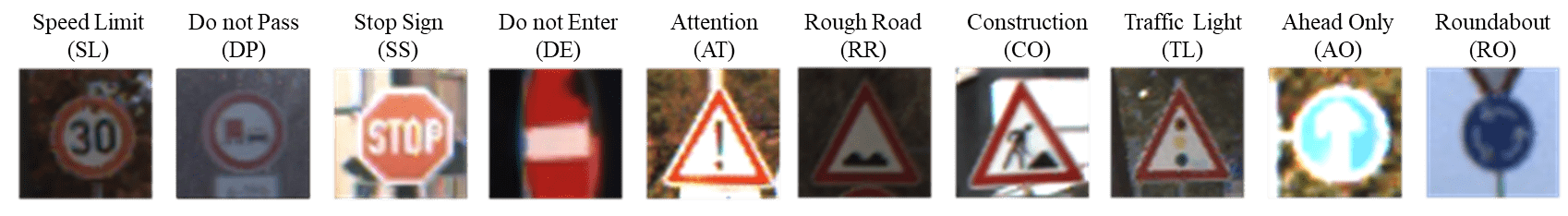}
	\caption{Ten traffic signs selected from the GTSRB dataset.}
    \label{fig:GTSRB_sample}
\end{figure*}

In this paper, we assume that an off-the-shelf detection algorithm
for sign detection is available, e.g., \cite{houben2013detection,zhu2016traffic,yang2015towards},
and it can be used to obtain a guess of the sign's ground truth label $\ell$
at every time $t\in [0, T]$.
For simplicity, we assume that input images contain only the signs
cropped from the camera's image.
Example images of the traffic signs are presented in Fig.~\ref{fig:GTSRB_sample}.
We use the German Traffic Sign Recognition Benchmark (GTSRB) dataset\footnote{Other datasets, e.g., \cite{neuhold2017mapillary}, can also be used with simple modifications.} to simulate the perception of the vehicle while driving toward the traffic sign.
We modify the images from the GTSRB dataset with time-varying resolution and pixel-wise noise, such that the observed image $\by$ is a random variable.
Examples of the modification are presented in~\S\ref{sec:modified_gtsrb}.

% In this paper, it is assumed that some well-known detection algorithms complete the traffic sign detection in advance of perceiving the label of the sign, e.g., \cite{houben2013detection,zhu2016traffic,yang2015towards}, and the perception model only takes detected traffic signs as input.
% The detected traffic sign images are taken from the German Traffic Sign Recognition Benchmark (GTSRB) dataset\footnote{The other datasets, e.g., \cite{neuhold2017mapillary}, can also be used with simple modification.}, some of the example images are presented in Fig. \ref{fig:GTSRB_sample}.
% To simulate scenario when the vehicle is driving toward the traffic sign,

\subsection{Perception Model} \label{sec:perception-model}

\begin{figure}[t]
    \centering
	\includegraphics[width=\linewidth]{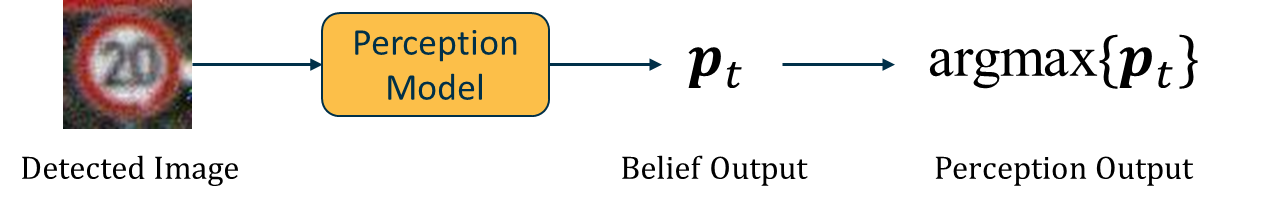}
	\caption{The traffic sign perception process.} 
    \label{fig:perception_model}
\end{figure}
% 	\nader{I would suggest not to call CNN a perception model. How about name it Learning Model? I would also call the $\arg\max$ as an Inference Unit. Should we put a Softmax function in between CNN and ${\bf p}_t$? } \guangyi{Yes, I'll fix it now. The value of argmax p is named as the perception output for the rest of the sections.}\nader{That's fine. We do not have much time. Let's go with what we have so far.}

The noisy observation $\by$ is fed into the perception model as an image. In this paper, we consider the perception model as a simple convolutional neural network (CNN) model, e.g., VGG-19 \cite{simonyan2014very}, such that $\bp = h(\by)$ can be rewritten as
\begin{equation}    \label{eq:perception}
    \bp  = \texttt{Softmax}\left(\,\text{CNN}(\by)\,\right),
\end{equation}
in which belief outputs $\bm{p}_t = (p_{t,1}, \cdots, p_{t,m}) \in \mathcal{P}_m$ are generated for all continuous time instances within $[0,T]$, as depicted in Fig. \ref{fig:perception_model}.

The focus of this work is not to improve the accuracy of detection/recognition algorithms but rather to integrate them with risk quantification to further strengthen the perception module against misperception by reasoning about the severity of the potential failure.
For clarity, we choose a simple detection model.
However, with simple modifications, the proposed approach applies to any detection model~\cite{haloi2015traffic,zhang2020lightweight}.

\subsection{Estimated Statistics of Belief Outputs}
To quantify the risk, one must obtain or estimate the statistics of the target random variable.
Measuring the statistics of belief output $\bp$ for the entire time-span $[0,T]$ is 
inefficient due to the time-varying resolution and noise in~\eqref{eq:modified_observation}.
To resolve this issue, we assume that the statistics of $\bp$ do not change drastically in any sufficiently short time interval $[t - \tau, t) \subset [0,T]$, $\tau \in \R_+$ and $t \in [\tau, T]$, and we can only obtain a finite number of observations in each time interval.
Let $\mathcal{T}_t^\tau \subset [0,T]$ be the finite set of (uniform or non-uniform) sampling times, and $q = |\mathcal{T}_t^\tau|$, the cardinality of the set $\mathcal{T}_t^\tau$.
Even for the interval $[t -\tau, t)$, quantifying the statistics of $\bp$ from the $q$ observations is not intuitive since the random variable follows the constraint
\begin{equation}    \label{eq:perception_output_constraint}
    \sum_i p_{t,i} = 1 \text{ and } p_{t,i} \geq 0 \text{ for all $i \in \mathcal{M}$},
\end{equation}
which falls within the belief space $\mathcal{P}_m$.
Unlike well-known distribution fitting approaches for Gaussian random variables, we estimate the statistics of $\bp$ using a Dirichlet distribution, for which the corresponding random variable satisfies~\eqref{eq:perception_output_constraint}.
A random variable $\bm{z} \in \mathcal{P}_m$ with the Dirichlet distribution $\mathcal{D}(\bm{z}, \bm{\alpha})$ has probability density function 
\begin{align} \label{eq:diri_density}
    f_{\mathcal{D}}(z_1,...,z_m; \alpha_1, ..., \alpha_m) = \frac{\Gamma(\sum_{i = 1}^{m} \alpha_i)}{\prod_{i = 1}^{m} \Gamma(\alpha_i)} \prod_{i = 1}^{m} z_i^{\alpha_i - 1},
\end{align}
where $\Gamma(\cdot)$ denotes the Gamma function, $\sum_{i \in \mathcal{M}} z_i = 1$, and $z_i \geq 0 $ for all $i \in \mathcal{M}$,
and $\bm{\alpha} \in \R^m_+$ is the concentration parameter vector of the $m$-order Dirichlet distribution.
% , which is a family of continuous multivariate probability distributions parameterized by a vector $\bm{\alpha} \in \R^m_+$

% 
% It is technically reasonable for us to obtain the statistics of $\bp$ for each sufficiently short time intervals, e.g.,  $[t -\tau, t)$ with $t - \tau, t \in [0,T]$ for some $\tau \in \R_+$. In this paper, we assume the belief outputs $\bp$ do not change drastically in time for some carefully selected $\tau$.
% }
% since for these short time intervals can be considered in the steady state with certain $\tau$ \footnote{This assumption is practical since most vehicles drive slow ($\sim 29$kmh, or $8$m/s \cite{bieker2018analysis}) toward the intersection.} since the speed of the vehicle is relatively slow.
%
% However, measuring the statistics of belief outputs for the entire time-span $[0,T]$ is feasible but inefficient since the outputs are considered not in a steady state given the time-varying resolution and noise level.

Let us consider $q$ images, sampled and processed over a time interval $[t-\tau, t)$, and their corresponding collection of belief outputs is given as a $q \times m$ matrix
% $
%     \bm{P}_{t} = [\bm{p}_{t - \tau},...,\bm{p}_{t}]^T.
% $
$
    \bm{P}_{t} = [\bm{p}_{t'}^T]_{t' \in \mathcal{T}^\tau_t}.
$
We use the fixed point approach proposed in~\cite{minka2000estimating} to estimate
the Dirichlet distribution, i.e., the concentration vector $\ba$, from a given set of belief outputs $\bm{P}_{t}$.
The method maximizes the log-likelihood of the estimated distribution and the original data.
Considering the convex nature of the problem \cite{ronning1989maximum}, the likelihood is unimodal, and its maximum can be obtained via a simple search \cite{minka2000estimating}.

% Then, the Dirichlet distribution of a given set of belief outputs $\bm{P}_{t}$ can be estimated using methods proposed in~\cite{minka2000estimating}, where the log-likelihood of the estimated distribution and the original data is maximized.
% Given the convex nature of the problem~\cite{ronning1989maximum}, the likelihood is unimodal, and its maximum can be obtained via a simple search.
% \guangyi{This paragraph needs a brief rewrite.}
% 
% In this paper, we use the fixed point approach proposed in~\cite{minka2000estimating} to estimate the value of the concentration vector $\ba$ from the given collection $\bm{P}_{t}$ over the time interval $[t-\tau, t)$.
% The estimated $\ba$ parameterizes the Dirichlet distribution $\mathcal{D}(\bz, \alpha_{t,1},...,\alpha_{t,m})$ for the random variable $\bz \in \mathcal{P}_m$, which accurately estimates the statistics of the perception outputs collected in $\bm{P}_{t}$.
Given an initial guess $\hat{\bm{\alpha}}_t$ of $\ba$, the estimated value of $\ba$ is updated using the following result. 

\begin{lemma}   \label{lem:diri_estimate}
    For a given set of belief outputs $\bm{P}_{t}$, there exists a set of values of $\ba$ which maximizes the log-likelihood, and $\ba$ can be updated element-wise using 
    \begin{equation}
        \Psi(\alpha_{t,i}^{new}) = \Psi(\sum_{j = 1}^{m} \alpha_{t,j}^{old}) + \frac{1}{q} \sum_{t' \in \mathcal{T}^\tau_t} \log p_{t',i},
    \end{equation}
    where $\Psi(\cdot)$ is the Digamma function,
    $\ba = (\alpha_{t,1},...,\alpha_{t,m})$,
    and $i \in \mathcal{M}$.
    % \guangyi{carefully review this}
\end{lemma}

The proof of the above result is omitted, and it can be obtained from \cite{minka2000estimating}. The above lemma provides the opportunity of estimating the statistics of belief outputs $\bp$ for the time interval $[t -\tau, t)$ with a Dirichlet random variable $\bz \sim \mathcal{D}(\bz, \ba)$, which allows us to compute the risk of misperception in a closed form.

%%%%%%%%%%%%%%%%%%%%%%%%%%%%%%%%%%%%%%%%%%%%%%%%%%%%%%%%%%%%%%%%%%%%%%%%%%%%%%%%%%%%%
\section{Risk of Misperception}   \label{sec:risk}

We introduce the cost for misperceiving traffic signs followed by our main result of misperception risk. 

\subsection{Cost of Traffic Signs Misperception}
\begin{figure}[t]
    \flushleft
    \hspace{24pt}
    \includegraphics[width=.05\linewidth]{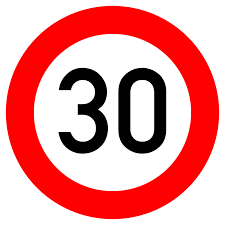}
    \hspace{5pt}
     \includegraphics[width=.05\linewidth]{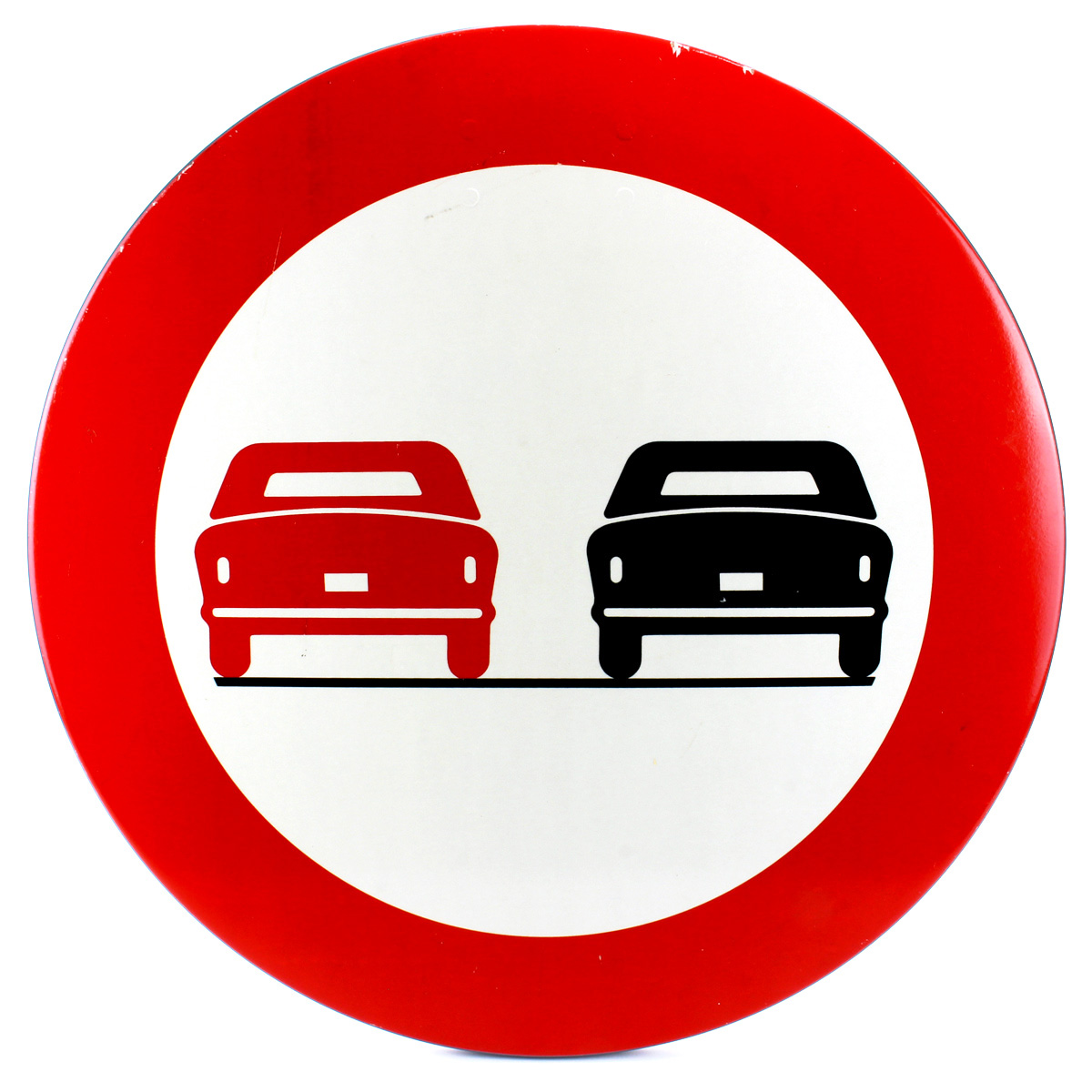} 
    \hspace{5pt}
     \includegraphics[width=.05\linewidth]{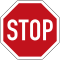} 
     \hspace{3pt}
     \includegraphics[width=.05\linewidth]{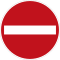} 
    \hspace{2pt}
     \includegraphics[width=.05\linewidth]{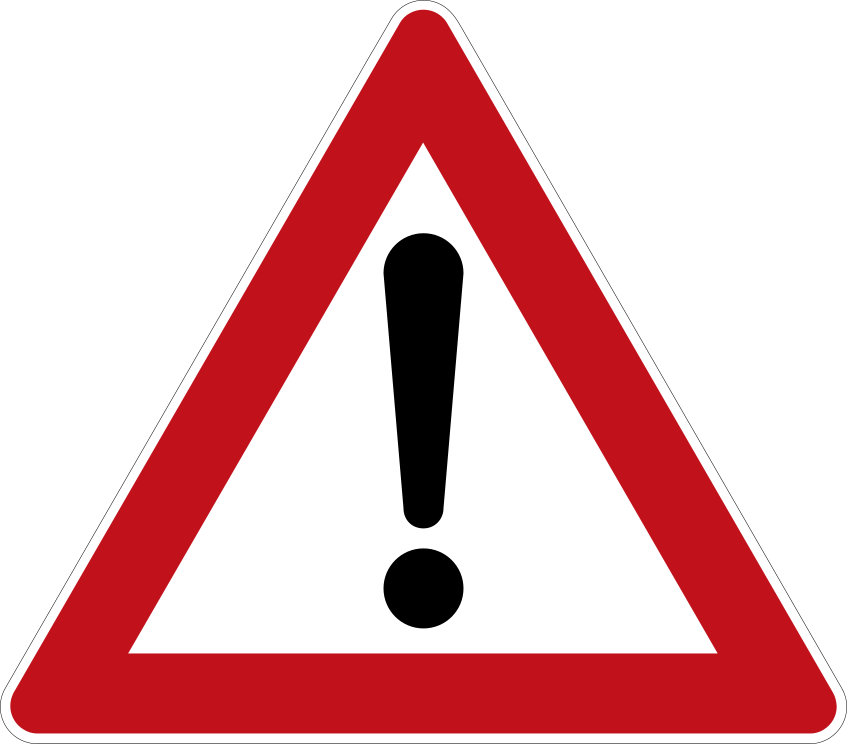} 
    \hspace{2pt}
     \includegraphics[width=.05\linewidth]{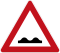}
     \hspace{2pt}
     \includegraphics[width=.05\linewidth]{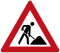} 
     \hspace{1.5pt}
     \includegraphics[width=.05\linewidth]{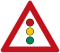} 
     \hspace{1pt}
     \includegraphics[width=.05\linewidth]{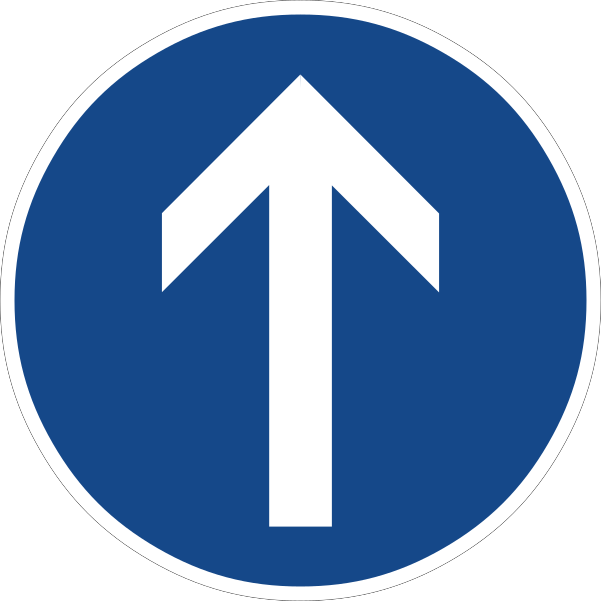} 
     \hspace{1pt}
     \includegraphics[width=.05\linewidth]{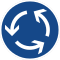} 
\end{figure}
\begin{table}[t]                           
	\centering
	\resizebox{\linewidth}{!}{ 
		\begin{tabular}{ l | l l l l l l l l l l l}
			\toprule
			Sign & SL & DP & SS & DE & AT & RR & CO & TL & AO & RO \\\hline
			SL  &0 & 174 & 103 & 103 & 123 & 123 & 121 & 103 & 121 & 120 \\
			DP &117 & 0 & 105 & 105 & 117 & 117 & 119 & 105 & 97 & 113 \\
			SS &135 & 109 & 0 & 96 & 110 & 110 & 110 & 96 & 135 & 135 \\
			DE &117 & 117 & 99.5 & 0 & 117 & 500 & 117 & 117 & 117 & 117 \\
			AT &71 & 111.5 & 102 & 92 & 0 & 50 & 0 & 102 & 51 & 137.5 \\
			RR &144.5 & 168 & 82 & 82 & 50 & 0 & 50 & 140 & 168 & 258 \\
			CO &102 & 41.5 & 82 & 82 & 30 & 0 & 0 & 41 & 83 & 173 \\
			TL &97 & 97 & 77.5 & 77.5 & 39 & 73 & 73 & 0 & 73 & 163 \\
			AO &91 & 91 & 86.5 & 86.5 & 45.5 & 45.5 & 45.5 & 91 & 0 & 182 \\
			RO &83 & 83 & 165 & 165 & 41.5 & 41.5 & 41.5 & 63 & 200 & 0 \\
			\bottomrule
	\end{tabular} }                        
	\caption{The cost metric for traffic sign misperception (unit: $\text{\texteuro} 1000$). }
	\label{table:cost}  
\end{table}
\vspace{-3pt}
Misperceiving traffic signs often leads to poor decisions of autonomous vehicles, which are primarily associated with high potential costs in real-world driving scenarios. 
Simply interpreting the belief output as ``correct'' or ``wrong'' does not provide adequate information for safe autonomous driving. 
The reason is that the high-level actions associated with each traffic sign do not yield the same potential cost, e.g., misperceiving the ``Speed-limit" sign as a ``Stop" sign induces a higher cost than misperceiving it as a ``Construction" sign. 

Therefore, to explore the severity of the misperception that is beyond correctness, we introduce a cost matrix of misperceiving the traffic signs, $C \in \R^{m \times m}$. For each label $i \in \mathcal{M}$, we define the cost of misperceiving the label $j \in \mathcal{M}$ as the label $i$ as $C_{ji}$.
The correct perception incurs zero cost, i.e., $C_{jj} = 0$.
The cost value is user-specific,
in this paper, we used the cost metric shown in Table~\ref{table:cost}, obtained by merging the estimated cost for the traffic sign-related violations~\cite{ayuso2010impact} for our selected traffic signs, such as potential fines and vehicle damages.

\subsection{Risk of Misperceiving Traffic Signs}
\begin{figure}[t]
    \centering
	\includegraphics[width=\linewidth]{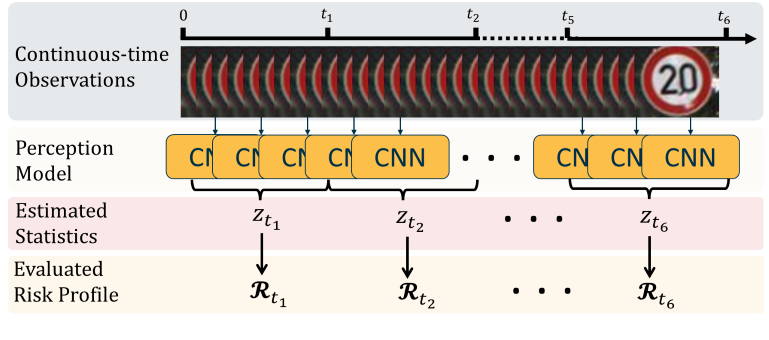}
	\caption{The above figure illustrates the evaluation of the risk profile for each short time interval, using $6$ time intervals as an example.}
    \label{fig:risk_model}
\end{figure}
The cost metric establishes the connection between the estimated belief output $\bz$ to the potential cost of misperception given by the cost metric. For each traffic sign, there exist $m-1$ possible cases that the estimated perception output $\arg\max \{\bz\}$ is incorrect. For a given estimated belief output $\bz$, one can deterministically define a family of nonlinear functions $r_i(\cdot)|_{i \in \mathcal{M}}$ that maps $\bz$ into the cost metric of misperceptions. For label $i$, the function $r_i(\cdot)$ takes value in $C_{ji}|_{j \in \mathcal{M}}$ such that 
\begin{equation}
    r_i(\bz) = C_{ji} \text{ if } \bz \in V_j.
\end{equation}
Then, the risk quantification can be adjusted to a family of discrete random variables, $r_i(\bz)$ for $i \in \mathcal{M}$.%, that takes value within the cost metric $C$. 

Given the fact that one or more instances of misperception may obtain the same cost value, let us denote the ordered cost vector as $\bm{c}_i \in \R^{m'_i}$, where the integer $m'_i \leq m$. The value of $m'_i$ denotes the number of unique values in all $C_{ji}|_{j \in \mathcal{M}}$, and the element of $\bm{c}_i$ obtains the unique values of $C_{ji}$ in a descending order such that
\begin{equation*}
    \max_{j \in \mathcal{M}} C_{ji} = (\bm{c}_i)_1 > \cdots > (\bm{c}_i)_{m'_i} = \min_{j \in \mathcal{M}} C_{ji}.
\end{equation*}

For the $i$'th label, there exist $m'_i$ possible cost values for $r_i(\bz)$. Then, the corresponding discrete probability distribution of $r_i(\bz)$ can be computed as follows.

\begin{lemma}   \label{lem:cost_prob}
    For each element of ordered cost vector $\bm{c}_i$, the probability of $\mathbb{P}\{r_i(\bz) = (\bm{c}_i)_j\}$ is given by
    \begin{equation}    \label{eq:r_prob}
        \mathbb{P}\{r_i(\bz) = (\bm{c}_i)_j\} = \hat{p}_{t,j} = \sum_{k|C_{k,i} = (\bm{c}_i)_j} \mathbb{P} \{\bz \in V_k\},
    \end{equation}
    where
    \begin{equation}    \label{eq:exceedance_prob}
        \mathbb{P} \{\bz \in V_k\} = \int_{0}^{\infty} \prod_{i \neq k} \left(\frac{\gamma(\alpha_{t,i},x)}{\Gamma(\alpha_{t,i})}\right) \frac{x^{\alpha_{t,k}-1}\exp{(-x)}}{\Gamma(\alpha_{t,k})} d x,
    \end{equation}
    and $\gamma(\alpha,x)$ is the lower incomplete gamma function.
\end{lemma}

The integral in \eqref{eq:exceedance_prob} can be obtained numerically using the approach proposed in \cite{soch2016exceedance} for computing the exceedance probability in the Dirichlet distribution. With the knowledge of the ordered cost vector $\bm{c}_i|_{i \in \mathcal{M}}$, the estimated statistics of belief outputs $\bz$, and its corresponding cost output $r_{i}(\bz)$, the Conditional Value-at-Risk of traffic sign misperception is shown in the following result.

\begin{theorem}     \label{thm:cvar}
    During the time interval $[t-\tau,t)$, given the estimated belief output $\bz$, the risk of misperception with the $i$'th label is given by
    \begin{equation}    \label{eq:thm-1}
        \mathcal{R}_{t,i}^{\varepsilon} = \frac{1}{\varepsilon} \left(\sum_{j = 1}^{v} (\bm{c}_i)_j \, \hat{p}_{t,j} + (\bm{c}_i)_{v+1} \big( \varepsilon - \sum_{j = 1}^{v} \hat{p}_{t,j}\big) \right),
    \end{equation}
    where the value of $v \in \Z$ is computed by
    \begin{equation}
        v = \sup_{v \leq m'_i} ~ \sum_{j = 1}^{v} \hat{p}_{t,j} \leq  \varepsilon,
    \end{equation}
    the value of $\hat{p}_{t,j}$ is obtained from \eqref{eq:r_prob}, and the value of $1 - \varepsilon$ represents the confidence level.
\end{theorem}

The above theorem provides the closed-form representation of the risk of misperception when the statistics of the belief output have a Dirichlet distribution. For a given confidence level $1-\varepsilon$, the magnitude of $\mathcal{R}^{\varepsilon}_{t,i}$ quantifies the severity of potential loss when perceiving the traffic sign as the $i$'th label based on the statistics of $\bz$. 

At each discrete time step $t$, let us also denote the collection of misperception risks for every label as the risk profile of misperception, such that
\begin{equation}    \label{eq:risk_profile}
    \bm{\mathcal{R}}_t^{\varepsilon} = [\mathcal{R}_{t,1}^{\varepsilon}, \mathcal{R}_{t,2}^{\varepsilon}, \cdots, \mathcal{R}_{t,m}^{\varepsilon}]^T \in \R^m,
\end{equation}
and the risk evaluation process is depicted in Fig. \ref{fig:risk_model}.
The \emph{risk output} is again a label obtained via the $\arg\min$ rule,
i.e., $\arg\min\{\bm{\mathcal{R}}_t^{\varepsilon}\} \in \mathcal{M}$.

\subsection{Accumulated Risk}
\begin{figure*}[t]
    \centering
	\includegraphics[width=\linewidth]{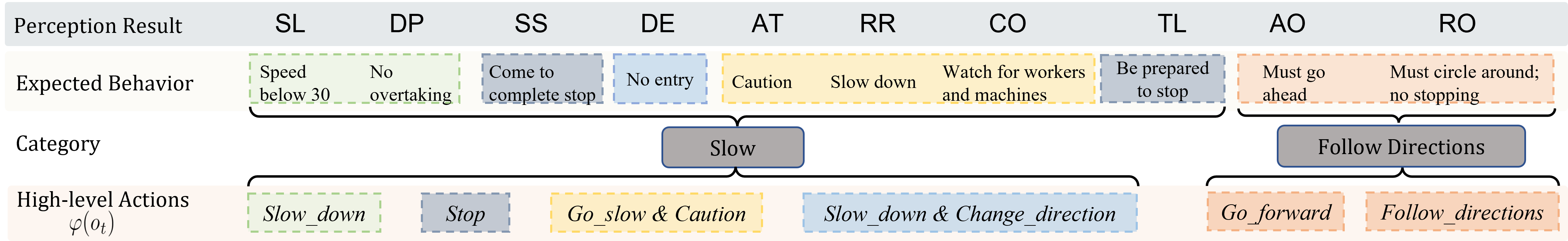}
	\caption{The high-level actions to be executed upon encountering a specific traffic sign.}
    \label{fig:control_actions}
\end{figure*}

The risk of misperception is capable of assessing the reliability of the belief output for a time interval $[t-\tau,t)$. However, in the real-world environment, the input quality is time-varying, and simply relying on the risk output for one short time does not reveal the truth about the target traffic sign. Thus, it also requires us to track the change of the risk throughout time and consider both current and past information. To this end, consider a weighted average of risk values with a scaling factor $\mu \in (0,1)$ which balances the importance of the present and the past risk values
%. The weighted sum of the past risk values is given by
% \begin{equation*}
%     S_{t,i} = \mu^{t-1} \mathcal{R}^{\varepsilon}_{1,i} + \mu^{t-2} \mathcal{R}^{\varepsilon}_{2,i} + ... + \mathcal{R}^{\varepsilon}_{t,i},
% \end{equation*}
% and its weighted average can be derived by computing the sum of the weights, i.e.,  $\sum_{k = 1}^{t} \mu^{t-k} = (1- \mu^{t})/(1-\mu)$, such that the normalized accumulated risk for perceiving as the $i$'th label is given by
% \cristi{
\begin{equation}    \label{eq:na_cvar}
    \hat{\mathcal{R}}_{t,i}^{\varepsilon} = \frac{1-\mu}{1- \mu^{K}} \sum_{k = 1}^{K} \mu^{K-k} \, \mathcal{R}_{k\tau,i}^{\varepsilon},
\end{equation}
where $t = K \cdot \tau$ is the current time at step $K \in \Z_+$,
$\mathcal{R}^{\varepsilon}_{k\tau,i}$ is evaluated at each time $t' = k\tau$ of step $k \in \{1, \ldots, K\}$ using~\eqref{eq:thm-1}, and $\mu \in (0,1)$ is the user-specified scaling factor. The corresponding risk profile can be stacked as $\bm{\hat{\mathcal{R}}}_{t}^{\varepsilon}$ using the same lines of argument in \eqref{eq:risk_profile}.
% The \emph{accumulated risk output} is $\arg\min\{\bm{\hat{\mathcal{R}}}_{t}^{\varepsilon}\}$.
% % }

The accumulated risk shows an evident advantage over the real-time risk $\mathcal{R}_{t,i}$ since it is more inclusive. Moreover, it can handle the situation when the observation changes drastically and the recent observations are unreliable since it does not solely rely on the most recent visual input. 
The accumulated risk is also normalized to enable tracking of the changes in the risk of misperception,
i.e., $\sum_{k = 1}^{K} \mu^{K-k} = (1- \mu^{K})/(1-\mu)$

% \subsection{Critical Risk Threshold} \label{sec:controller}
% \begin{figure}[t]
%     \centering
% 	\includegraphics[width=\columnwidth, scale= 0.98]{figs/control_categories.png}
% 	\caption{Categories of predefined control actions. \guangyi{This is going to be a table.}}
%     \label{fig:control_categories}
% \end{figure}

Let us also introduce the critical risk threshold $\eta \in \R_+$ as the user-defined maximum acceptable cost of the system.
It can be appropriately designed by carefully considering the task and the environmental factors.
Once the accumulated risk value $\hat{\mathcal{R}}_{t,i}^{\varepsilon}$ becomes lower than $\eta$, the corresponding {\it accumulated risk output} $o_t \in \mathcal{M}$ is considered as the label,
i.e., it is the minimal element in the risk profile such that
\begin{equation}
    o_t = {\arg\min}_{i \in \mathcal{M}} \{\hat{\bm{\mathcal{R}}}_t^{\varepsilon}\}, \text{ and } \hat{\mathcal{R}}_{t, o_t}^{\varepsilon} \leq \eta.
\end{equation}
% where the critical risk threshold $\eta$ is a user-specified parameter, and it can be appropriately designed by carefully considering the task and the environmental factors. 

% %%%%%%%%%%%%%%%%%%%%%%%%%%%%%%%%%%%%%%%%%%%%%%%%%%%%%%%%%%%%%%%%%%%%%%%%%%%%%%%%%%%%%
\section{Case Study}    \label{sec:case}
\begin{figure*}[t]
    \begin{subfigure}[t]{.24\linewidth}
        \centering
    	\includegraphics[width=\linewidth]{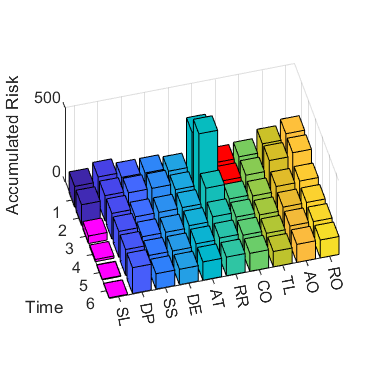}
    	\caption{True label: Speed-limit (SL) \\ \centering $\varphi(SL) =$\textit{Slow\textunderscore down}}
    \end{subfigure}
    \hfill
    \begin{subfigure}[t]{.24\linewidth}
        \centering
    	\includegraphics[width=\linewidth]{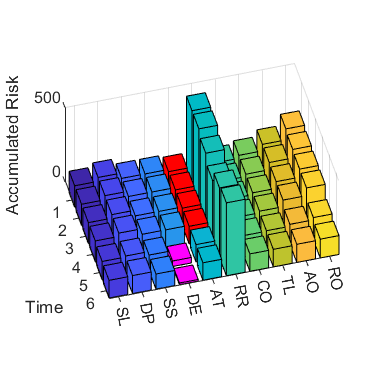}
    	\caption{True label: Do not enter (DE)\\ \centering $\varphi(DE) =$\textit{Slow\textunderscore down \& Change\textunderscore direction}}
    \end{subfigure}
    \hfill
    \begin{subfigure}[t]{.24\linewidth}
        \centering
    	\includegraphics[width=\linewidth]{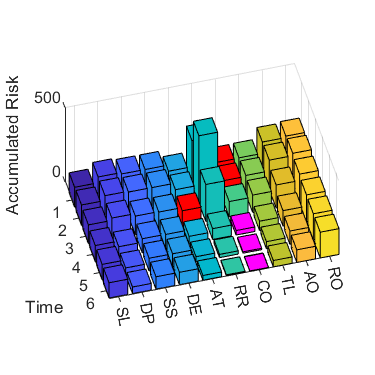}
    	\caption{True label: Construction (CO) \\ \centering $\varphi(CO) =$\textit{Go\textunderscore slow \& Caution}}
            \label{fig:traffic_risk_co}
    \end{subfigure}
    \hfill
    \begin{subfigure}[t]{.24\linewidth}
        \centering
    	\includegraphics[width=\linewidth]{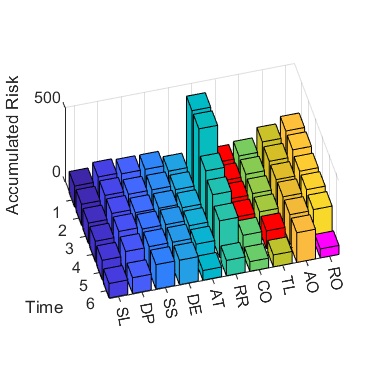}
    	\caption{True label: Roundabout (RO)\\ \centering $\varphi(RO) =$\textit{Follow\textunderscore directions}}
            \label{fig:traffic_risk_ro}
    \end{subfigure}
    \caption{The accumulated risk profile of misperception for various traffic signs $\bm{\hat{\mathcal{R}}}_{t}^{\varepsilon}$. At each time step, the label corresponding to minimum risk is shown in red, and the bar indicating $o_t$ is shown in magenta.}
    % \disha{The subcaptions indicate the true perception label and chosen high-level actions. }}
    \label{fig:traffic_risk}
\end{figure*}

In this case study, the perception model is trained with the original GTSRB dataset, and the risk of misperception is evaluated with the modified image. Let us consider $\tau = \frac{T}{6}$ and split $[0,T]$ in to $6$ intervals for all case studies. We now define some notions used for evaluating the proposed risk metric and then proceed with a detailed discussion on simulations.

%%%%%%%%%%%%%%%%%%%%%%%%%%%%%%%%%%%%
\subsection{High-Level Actions and Time To Execution}
% \noindent \textbf{Available Control Actions: }
% \disha{TODO: 
% \begin{align*}
%     \text{max} \; \; t_{exec}\\ 
%     \text{where}, t_{exec} =(T - t) . I (\mathcal{R}_t \leq \eta)
% \end{align*}
% }
\iffalse
\begin{table}[t]

% \caption{Environment and System Variables}
% \vspace{-3 pt}
\centering
\begin{tabular}{|p{0.14\linewidth} | p{0.19\linewidth}| p{0.22\linewidth}| p{0.22\linewidth}|}

\hline
 Mandatory & Performance & Directional & Precautionary\\
\hline\hline
\textit{Stop} & \textit{Constant\textunderscore{speed}}
 & \textit{Change\textunderscore{direction}}& \textit{Caution}\\ 
 \textit{Slow\textunderscore{down}} &  & \textit{Move\textunderscore{forward}} & \textit{Headlights\textunderscore{on/off}}\\ 
  &   & \textit{Roundabout} &  \\\hline
    % \textit{} & \textit{} & \textit{} & \textit{}\\

\end{tabular}
\caption{Categories of available high-level actions}
\label{Tab:actions}
\vspace{-5pt}
\end{table}
\fi

We consider a set $\Lambda$ of prescribed high-level actions that allow the system to execute an appropriate maneuver corresponding to the detected sign as depicted in Fig. \ref{fig:control_actions}. The high-level actions are the abstractions of actuation commands to the system.
 
%  Since the focus of this work is to strengthen the perception result by quantifying the risk associated with misperception, we associate each identified minimum risk label $o_t$ to a set of 
% \iffalse
% We categorize the available actions into \textit{ mandatory\textunderscore actions, performance\textunderscore actions, directional\textunderscore actions} as shown in Table~\ref{Tab:actions}. Mandatory and performance actions result in mutually conflicting actuation commands and hence cannot be chosen together. Along with these actions, the system is capable of executing an additional set of \textit{precautionary\textunderscore actions} that comprise actions such as \textit{caution} and \textit{headlights\textunderscore on}, that do not result in physical movement of the system \cristi{\sout{in the workspace $\Im$}} but still may be necessary for negotiating the given traffic situation such as navigating through a work zone or a tunnel.
% \fi
Given the accumulated risk output $o_t$,  $\varphi : \mathcal{M} \rightarrow \Lambda$ provides an action $\lambda \in \Lambda$ to be executed.
In order to facilitate the execution of the chosen high-level actions, our framework maximizes the \textit{time to execution} ($t_{exec}$) defined as follows: 
\begin{align}
    t_{exec} =(T - t) \cdot I (\hat{\bm{\mathcal{R}}}_t^{\varepsilon} \leq \eta),
\end{align} 
where $I(\cdot)$ is an indicator function and takes the value 1 whenever $\hat{\bm{\mathcal{R}}}_t^{\varepsilon}$ drops below the risk threshold $\eta$ and is 0 otherwise. Thus, once the normalized accumulated risk crosses the acceptable cost $\eta$, a corresponding high-level action can be chosen.

As this work focuses on strengthening perception-related safety using risk quantification, we consider a simple mapping between the accumulated risk output $o_t$ to the action space $\Lambda$.
Developing risk-aware controllers is a topic for future investigation.

% Fig.~\ref{fig:control_actions} shows the high-level action primitives that need to be executed by the vehicle. Note that, even in case of misperception, some cases lead to a lower risk as compared to others. e.g., 

\subsection{Modified GTSRB Dataset} \label{sec:modified_gtsrb}
\begin{figure}[t]
    \centering
	\includegraphics[width=\linewidth]{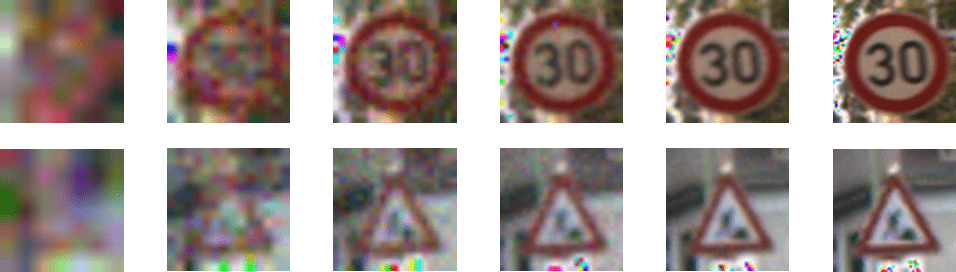}
	\caption{Example images from the modified GTSRB dataset. Shown for $6$ time intervals.}
    \label{fig:speed_example}
\end{figure}
Since the images from the GTSRB dataset are static and do not contain any specific types of noise, we apply the following modifications to the dataset, which simulates the scenarios of the vehicle approaching the traffic sign:
\begin{itemize}
    \item The quality of the detected image is time-varying. The resolution of the detected images increases when the vehicle gets closer to the traffic sign, i.e., as $t$ increases. 
    \item The independent time-varying Gaussian noise is added to each pixel of the image. The magnitude of the noise, $b_t$, decreases as the vehicle gets closer to the traffic sign, i.e., as $t$ increases.  
\end{itemize}

We select $10$ types of traffic signs from the dataset, and examples of the modified image are shown in Fig. \ref{fig:speed_example}. The above modification can be represented using \eqref{eq:modified_observation}, in which we consider $g(\cdot)$ is the function that changes the resolution w.r.t the time $t$, and $b_t$ is the time-varying noise magnitude. For instance, our choices are $\dim(g(t, \bm{y}_0)) = \frac{t}{T} \dim(\bm{y}_0)$ and $b_t = 0.02 \frac{T}{t}$.

\subsection{Simulations}

\begin{figure}[t]
    \begin{subfigure}[t]{.32\linewidth}
        \centering
    	\includegraphics[width=\linewidth, scale =0.2]{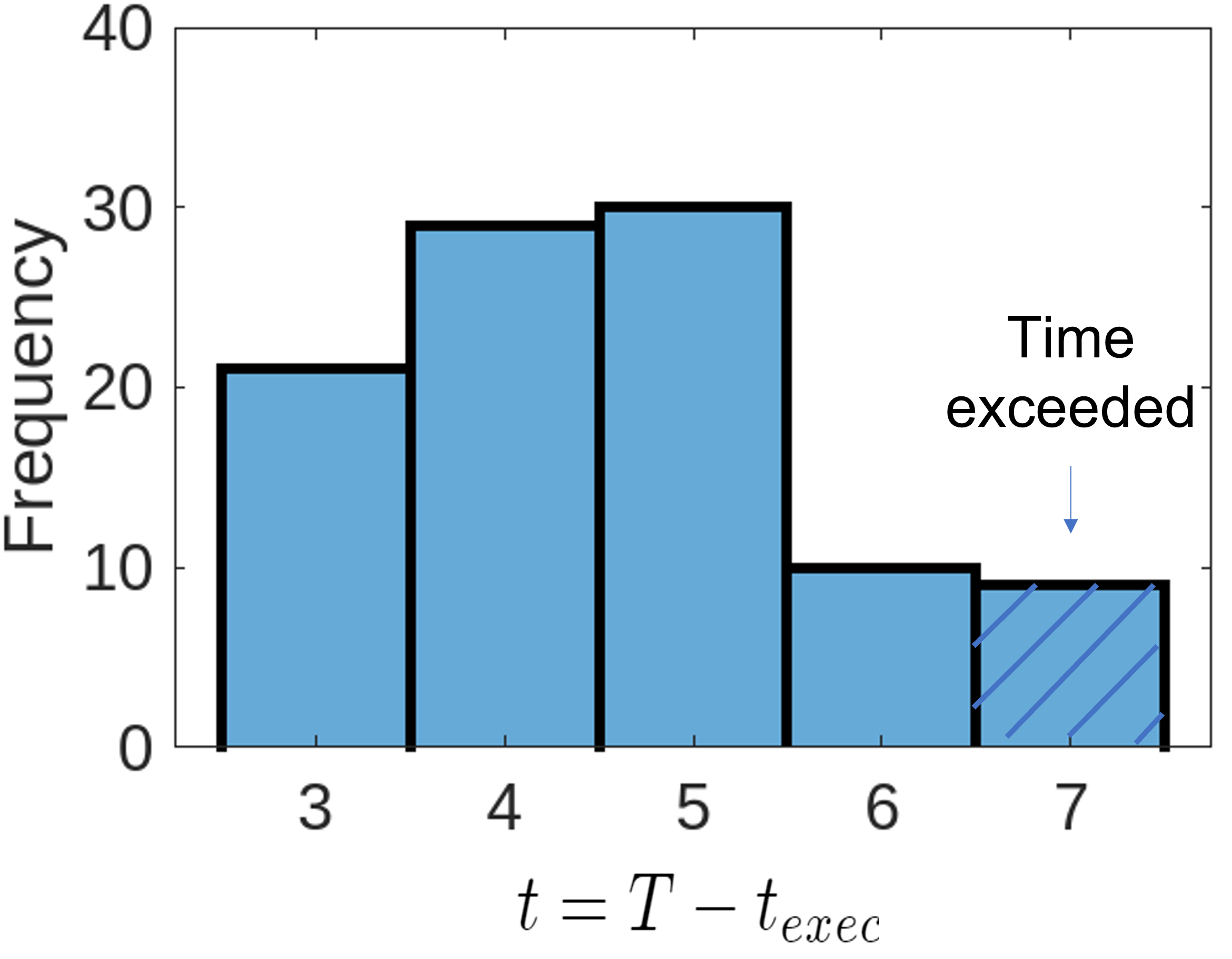}
    	\caption{$\eta = \text{\texteuro} 1000$.}
    \end{subfigure}
    \hfill
    \begin{subfigure}[t]{.32\linewidth}
        \centering
    	\includegraphics[width=\linewidth]{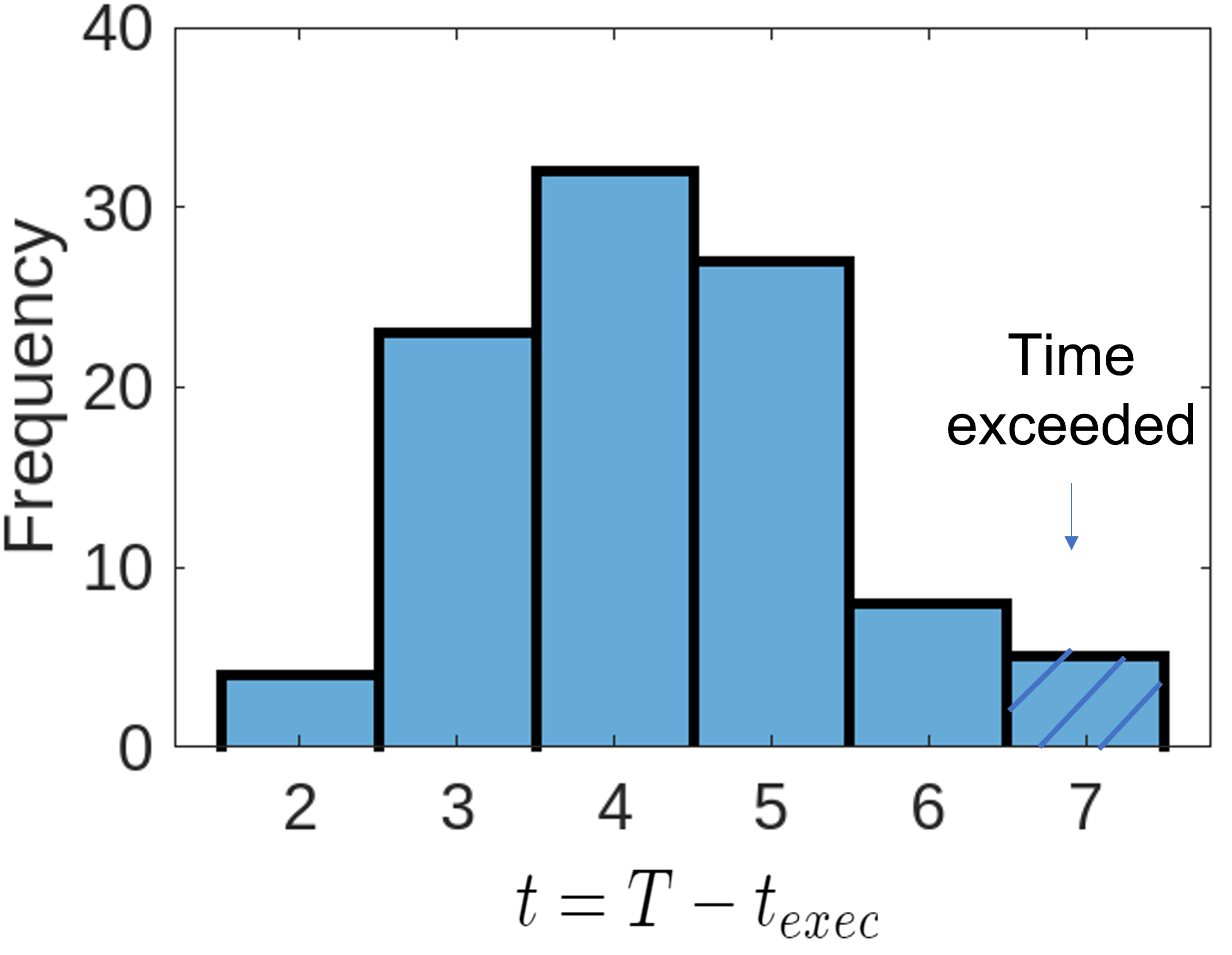}
    	\caption{$\eta = \text{\texteuro} 10000$.}
    \end{subfigure}
    \begin{subfigure}[t]{.32\linewidth}
        \centering
    	\includegraphics[width=\linewidth]{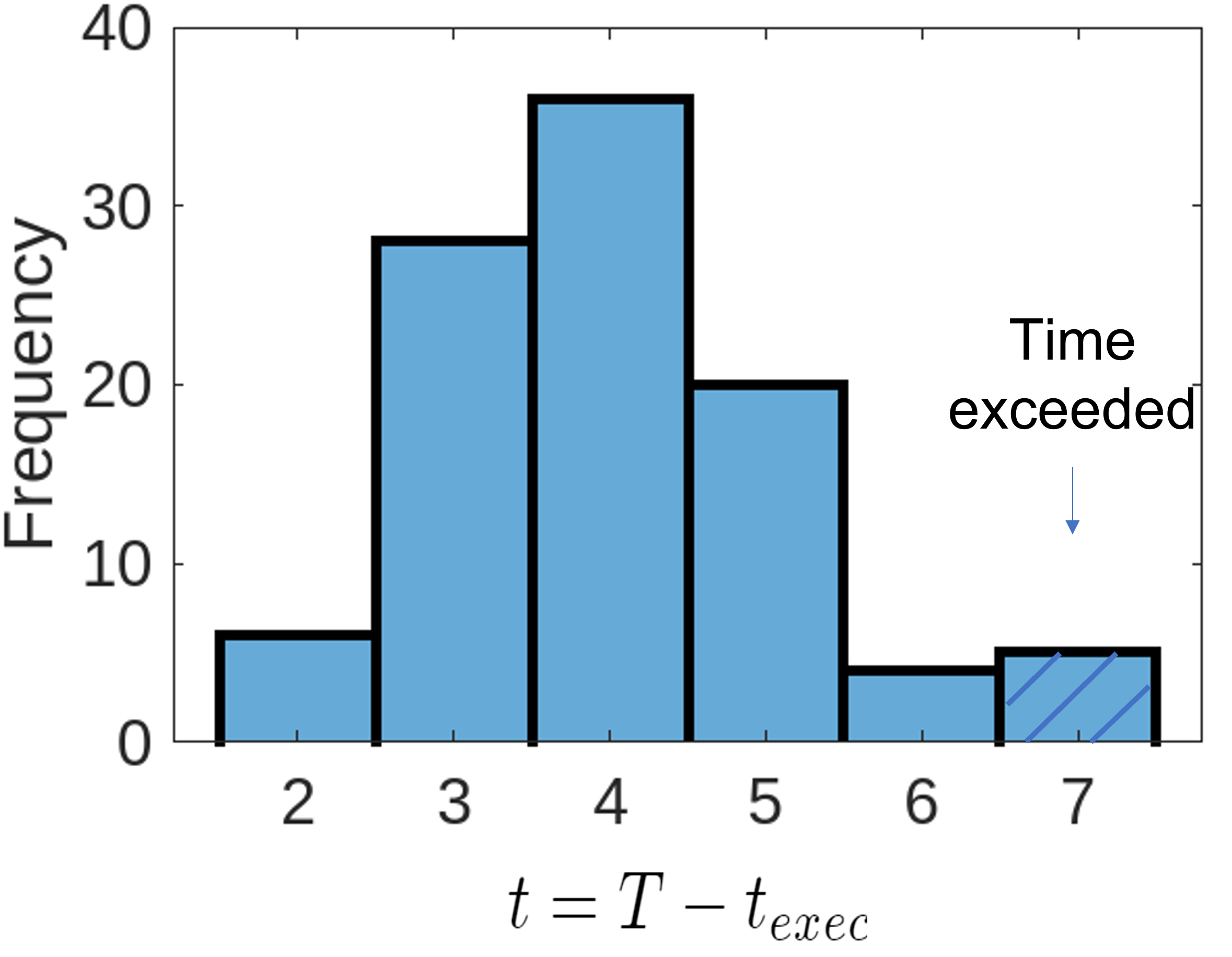}
    	\caption{$\eta = \text{\texteuro} 50000$.}
    \end{subfigure}
    \caption{The distribution of $T - t_{exec}$ with various $\eta$. Evaluated among $100$ trails.}
    \label{fig:time_test}
\end{figure}

\subsubsection{Risk of Traffic Sign Misperception}
Using the result from Theorem \ref{thm:cvar}, we evaluate the risk of misperceiving specific signs, for which some examples are shown in Fig. \ref{fig:traffic_risk}. The tested images are taken from the modified GTSRB dataset with $\eta = \text{\texteuro} 10000$ and the confidence level is set by letting $\varepsilon = 0.1$. Among all presented cases, the ground truth labels of the sign $\ell \in \mathcal{M}$ are associated with the accumulated risk output $o_t$ by $t = 6$. 
% \disha{At each time step, the label corresponding to minimum risk is shown in red and magenta before and after crossing the critical risk threshold. The first instance of magenta is when an action decision can be made with an acceptable cost. } \guangyi{I moved this part to the caption.}

There are a few interesting observations that are worth reporting: The label ${\arg\min}_{i \in \mathcal{M}} \{\bm{\hat{\mathcal{R}}}_{t}^{\varepsilon}\}$ commonly occurs at ``CO" when the input image is noisy and with low resolution. This phenomena is because the action related to ``CO" is ``\textit{Go\textunderscore slow \& Caution}" (see Fig. \ref{fig:control_actions}), which is intuitively the safest action when the visual input is not reliable. The severity of different instances of misperceptions depends on the expected high-level actions to be executed, e.g., in Fig. \ref{fig:traffic_risk_co}, the high-level actions associated with ``AT", ``RR", and ``CO" are the same, and thus, their corresponding accumulated risks lie within the same range of values. Therefore, the system will not incur a penalty even in case of misperception. In Fig. \ref{fig:traffic_risk_ro}, the risk output $o_t$ is only obtained at $t = 6$, since the potential loss of misperceiving the label ``RO" is relatively high, and the decision should only be made when the visual input is reliable enough, i.e., $\hat{\mathcal{R}}_{t, o_t}^{\varepsilon} \leq \eta$.

\subsubsection{Critical Risk Threshold}
Given different values of the critical risk threshold $\eta$, it is expected to observe various distributions of the time to execution $t_{exec}$, which is presented in Fig. \ref{fig:time_test}. The result indicates that with the increase of the threshold $\eta$, the autonomous vehicle could make earlier decisions owing to the higher acceptable potential cost. Thus, the distribution of $t_{exec}$ is more concentrated in higher values, or the distribution of $T - t_{exec}$ is more concentrated toward lower values.
Our result also reveals a potential trade-off between $\eta$ and $t_{exec}$ as a smaller choice of $\eta$ may prevent autonomous vehicles from making early decisions.

\subsubsection{Risk Outputs vs Perception Outputs}
\begin{figure}[t]
    \begin{subfigure}[t]{\linewidth}
        \centering
    	\includegraphics[width=\linewidth]{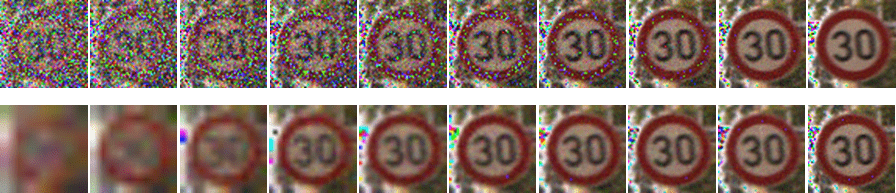}
    	\caption{Example images with: 1) Fixed relative resolution ($100\%$) with various noise levels; 2) Fixed noise level $b_t = 0.04$ with various resolution levels;}
    \end{subfigure}
    \medskip
    \begin{subfigure}[t]{\linewidth}
        \centering
    	\includegraphics[width=\linewidth]{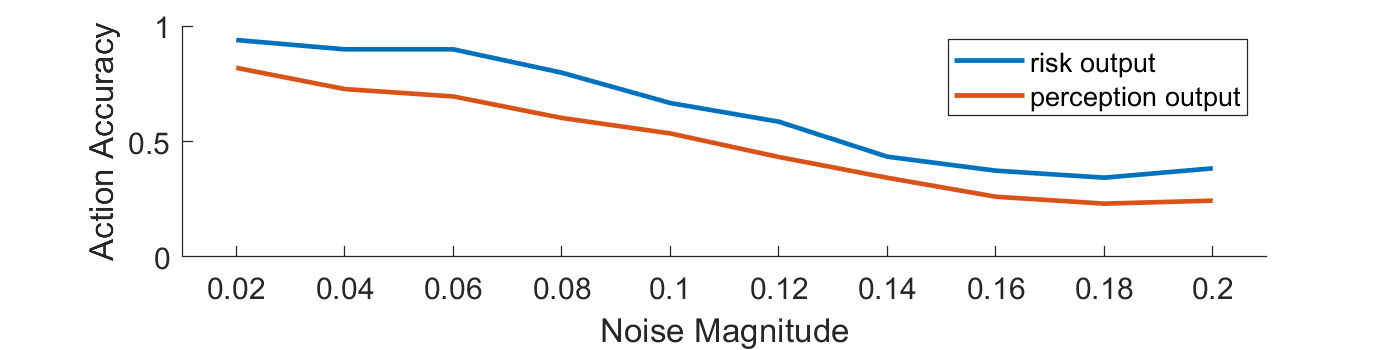}
    	\caption{Action Accuracy vs Noise.}
    \end{subfigure}
    \hfill
    \begin{subfigure}[t]{\linewidth}
        \centering
    	\includegraphics[width=\linewidth]{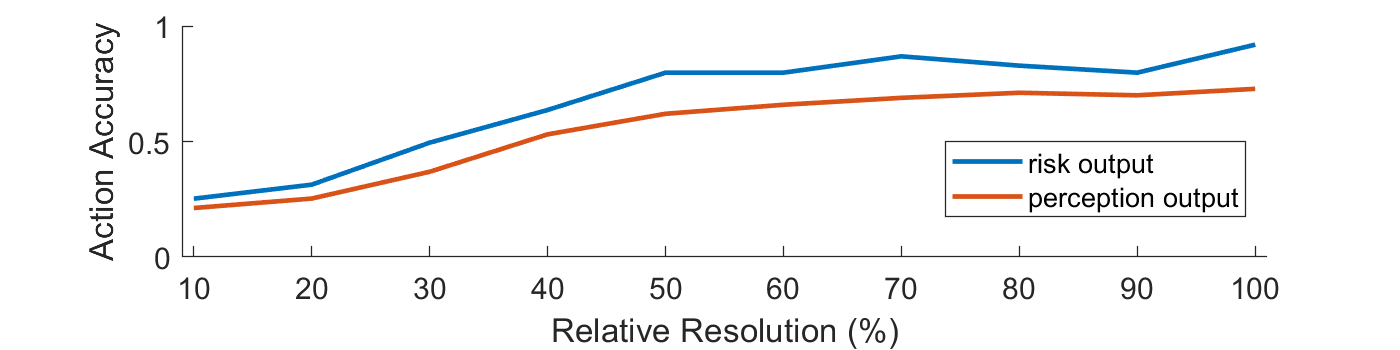}
    	\caption{Action Accuracy vs Resolution.}
    \end{subfigure}
    \caption{Action accuracy with various noise levels and resolutions.}
    \label{fig:noise+resolution}
\end{figure}

The risk output ${\arg\min}_{i \in \mathcal{M}} \{\bm{\mathcal{R}}_t^{\varepsilon}\}$ exhibits a significant advantage compared to the perception output $\arg \max \{\bp\}$ in the view of preventing the potential losses\footnote{The result is validated for a fixed time interval without using accumulated risk to obtain the independent performance for each noise and resolution level.}. We compare the ratio of correct high-level actions, i.e., the action accuracy, generated with both approaches with various noise and resolution levels, as shown in Fig. \ref{fig:noise+resolution}. In both cases, the risk output ${\arg\min}_{i \in \mathcal{M}} \{\bm{\mathcal{R}}_t^{\varepsilon}\}$ outperforms the perception output $\arg \max \{\bp\}$ by nearly $20\%$ in the action accuracy. It is because the consequences of each high-level action are inclusively considered in the risk of misperception, which exhibits the major difference between these two approaches.
% The expectation of $r_{t,i}$, i.e., $\mathbb{E}[r_{t,i}]$, also suggests the correct label by considering the statistics of $\bp$. However, it does not capture the severity of the potential failure, which is revealed by the risk of misperception, and it may underestimate the consequences of the decision.

% \subsubsection{Risk of Misperception with Various Noise and Resolutions Levels}

\section{Conclusion} \label{sec:conclusion}
In this work, we address the problem of mitigating the effects of misperceiving traffic signs for autonomous driving given noisy visual input. Using the well-known CVaR measure, we construct the framework that evaluates the risk of misperception as a function of the estimated belief output statistics and the user-specified cost metric that captures the severity of potential failures to the system in the event of misperception. Furthermore, leveraging the gradual improvements in the detection accuracy due to gradually improving sensing resolution and smaller effect of noise, we define a discounted accumulated CVaR-based risk. The proposed risk measure reveals the risk output and the accumulated risk output that can be utilized for decision-making and control with any controller of choice. The extensive case studies demonstrate the effectiveness of the proposed risk-quantification framework.

This work is the first step in incorporating the risk of misperception into the perception-planning framework with a focus on the perception module. The immediate natural extension of the work is to design a risk-aware controller to guarantee desirable properties such as system safety and minimum time execution. It also enables the analysis of the perception model from the view of the misperception risk when the output statistics can be explicitly written as a function of model parameters~\cite{amini2021robust}.

\printbibliography

\end{document}